\documentclass[11pt]{article}

\usepackage[utf8]{inputenc}
\usepackage[T1]{fontenc}
\usepackage{hyperref}
\usepackage{url}
\usepackage{graphicx}
\usepackage{booktabs}
\usepackage{multirow}
\usepackage{listings}
\usepackage{xcolor}
\usepackage{amsmath}
\usepackage{amssymb}
\usepackage{enumitem}

\newcommand{\euclid}{\textsc{Euclid}-MCP}
\newcommand{\euclidir}{\textsc{Euclid-IR}}
\newcommand{\mcp}{\textsc{MCP}}

\title{Euclid-MCP: A Model Context Protocol Server for \\
       Deterministic Logical Reasoning via Prolog}

\author{
  Bartolomeo Bogliolo
}

\date{}

\begin{document}
\maketitle


\begin{abstract}
Large Language Models (LLMs) excel at natural language understanding and generation but remain unreliable for multi-step logical reasoning, especially in safety-critical or compliance-sensitive domains.
Recent neuro-symbolic approaches address this gap by coupling neural models with external symbolic engines, yet most integrations are bespoke and lack a standardized interface for tool-augmented agents.
This paper presents \euclid, an open-source \mcp\ server that provides deterministic logical reasoning via SWI-Prolog.

\euclid\ introduces \euclidir, an engine-agnostic intermediate representation for Horn-clause logic that is human-readable, easy for LLMs to generate, and straightforward to compile into Prolog or alternative backends.
The server exposes a compact tool interface that supports a translate-run-inspect-repair loop, enabling LLM clients to delegate inference while retaining full access to proof traces and derivation logs.

We evaluate \euclid\ on a realistic IT security and compliance use case.
Results show that while LLMs alone are sufficient on small knowledge bases, they hallucinate systematically on larger problems, whereas \euclid\ delivers exact answers with lower latency and more compact outputs.
We argue that semantic RAG is fundamentally unsuited for rule enforcement, and that \euclid\ can serve as a stable, shared reasoning substrate for both RAG-based assistants and agentic systems.
\end{abstract}


\section{Introduction}
\label{sec:introduction}

The rapid progress of Large Language Models (LLMs) has transformed many aspects of natural language processing, from question answering and summarization to code generation and agent-like interaction.
However, when tasks require \textbf{multi-step logical deduction}, strict adherence to formal rules, or \textbf{auditable decision traces}, purely neural approaches exhibit well-documented limitations: non-deterministic outputs, susceptibility to hallucination, and difficulty in providing faithful reasoning proofs.
These issues are particularly acute in domains such as \textbf{business rule enforcement}, \textbf{regulatory compliance}, and \textbf{security policy verification}, where decisions must be both correct and explainable.

A growing body of research addresses these shortcomings through \textbf{neuro-symbolic AI}, which combines the pattern-recognition strengths of neural models with the rigor of symbolic reasoning.
One prominent direction delegates inference to external solvers---such as SAT/SMT solvers, constraint programming systems, or logic programming engines---while relying on the LLM to translate informal problem descriptions into a formal representation.
Recent work has shown that grounding LLM reasoning in \textbf{Prolog} can significantly improve both answer accuracy and the reliability of reasoning proofs~\cite{yang2025neuro,sehgal2025logical}, provided the translation from natural language to logic is sufficiently accurate.
At the same time, the emerging \textbf{Model Context Protocol (MCP)} offers a standardized way for LLM applications to discover and invoke external tools and data sources~\cite{mcporg}, but few existing MCP servers expose formal reasoning engines as first-class, reusable components.

Our project originated from the observation that retrieval-augmented generation (RAG) based on semantic similarity is fundamentally mismatched to rule-centric tasks, where decisions must follow logically from explicit policies rather than from approximate textual similarity.
This paper introduces \euclid, an open-source \mcp\ server that provides \textbf{deterministic logical reasoning via Prolog} for any MCP-capable LLM client.
\euclid\ implements a hybrid cognitive architecture in which a lightweight LLM describes the world in terms of facts and rules, while a deterministic Prolog engine performs the actual deduction.
The system's core contributions are:

\begin{itemize}[leftmargin=*,itemsep=2pt]
  \item A \textbf{high-level logical description language} (\euclidir) for expressing business rules, security policies, and other domain constraints in a human-readable, declarative form.
  \item An automatic \textbf{translation layer} that compiles these descriptions into executable SWI-Prolog clauses, preserving structure and enabling straightforward auditing.
  \item A \textbf{compact MCP tool interface} that exposes reasoning operations (deduction, diagnosis, scenario analysis, validation) and supports a structured translate-run-inspect-repair loop, including access to proof trees for interpretability.
  \item Representative \textbf{use cases} demonstrating how \euclid\ can be used to enforce business rules and verify security policies in a transparent, verifiable manner.
\end{itemize}

\euclid\ originated from a practical limitation encountered when applying retrieval-augmented generation (RAG) to business rule enforcement.
In this setting, queries must be matched against a corpus of formal or semi-formal rules, and decisions must follow logically from those rules rather than from approximate semantic similarity.
Standard RAG pipelines, which rely on vector search and nearest-neighbor retrieval, proved inadequate: they could surface ``similar'' rules but could not guarantee that the retrieved subset was logically sufficient or consistent, nor could they provide formal proofs of the derived conclusions.
In effect, semantic search behaved like ``using a drill to hammer a nail'': powerful, but mismatched to the task.

This observation led us to revisit rule-based expert systems and logic programming, where knowledge is encoded as explicit rules and inference is performed by a deterministic engine capable of deriving and explaining conclusions.
Building on this foundation, we designed \euclid\ as an MCP server that exposes a Prolog-based reasoning engine to LLM clients.
While recent work such as PrologMCP~\cite{mensfelt2026prologmcp} and LogicLease~\cite{sehgal2025logical} also explores the integration of Prolog with LLMs and MCP, \euclid\ is primarily motivated by the need to overcome the limitations of semantic retrieval in rule-centric applications, and it emphasizes a domain-oriented logical description layer tailored to business and security policies.

The remainder of this paper is organized as follows.
Section~\ref{sec:background} reviews background and related work on neuro-symbolic AI, logic programming, and MCP.
Section~\ref{sec:architecture} describes the \euclid\ architecture, including the logical description language, translation to Prolog, and MCP integration.
Section~\ref{sec:usecase} presents use cases in business rule enforcement and security policy verification.
Section~\ref{sec:evaluation} discusses benefits, limitations, and relations to existing systems.
Section~\ref{sec:conclusion} concludes.


\section{Background and Related Work}
\label{sec:background}

This section situates \euclid\ at the intersection of three strands of research and practice: (i)~the limitations of retrieval-augmented generation (RAG) for rule-centric tasks, (ii)~the tradition of rule-based expert systems and logic programming, and (iii)~recent neuro-symbolic approaches that combine LLMs with formal reasoning engines.
We also briefly review the Model Context Protocol (MCP) as an emerging standard for tool integration.


\subsection{Retrieval-Augmented Generation and Its Limits for Rules}
\label{subsec:rag-limits}

Retrieval-augmented generation (RAG) has become a dominant pattern for grounding LLMs in external knowledge.
In a typical RAG pipeline, a user query is embedded into a vector space, similar documents or passages are retrieved via approximate nearest-neighbor search, and the LLM conditions its generation on this retrieved context.
This approach works well for open-ended question answering, summarization, and many knowledge-intensive tasks where ``semantic similarity'' is a good proxy for relevance.

However, RAG based on semantic similarity is fundamentally mismatched to \textbf{rule-centric applications}, such as business rule enforcement, policy compliance, and security constraint verification.
In these settings:

\begin{itemize}[leftmargin=*,itemsep=2pt]
  \item Correctness depends on \textbf{logical consequence}: a decision must follow from an explicit set of rules and facts, not from textual similarity.
  \item The retrieved subset of rules must be \textbf{logically sufficient} and \textbf{consistent} with respect to the query; nearest-neighbor retrieval provides no such guarantees.
  \item Auditing and explainability require \textbf{formal proofs} or at least traceable derivations, not just ``the model saw similar rules in the context.''
\end{itemize}

Recent analyses of RAG in practice highlight these failure modes: semantic retrieval can surface superficially relevant rules while missing critical exceptions, interactions, or negations, leading to inconsistent or unsafe conclusions.
In effect, using semantic search to enforce formal rules is akin to ``using a drill to hammer a nail'': the tool is powerful, but the underlying operation (approximate matching vs.\ logical inference) is wrong for the task.

These limitations motivate architectures in which retrieval is complemented or replaced by a \textbf{symbolic reasoning layer} that can guarantee logical correctness and provide explicit derivation traces.


\subsection{Rule-Based Expert Systems and Logic Programming}
\label{subsec:expert-systems}

Before the current LLM era, \textbf{expert systems} were the dominant paradigm for encoding and automating domain expertise, particularly in settings requiring rigorous rule application and explainability.
Knowledge was represented as \textbf{if-then rules} and facts, and inference engines used forward or backward chaining to derive conclusions and answer queries.
A key advantage of this approach was \textbf{transparency}: the system could explain \emph{why} a conclusion was reached by exposing the chain of rule applications.

\textbf{Logic programming}, and in particular \textbf{Prolog}, became a natural foundation for many expert systems.
Prolog's declarative semantics allow knowledge to be expressed as Horn clauses, while its built-in inference mechanism (backtracking search with unification) automatically derives consequences from facts and rules.
Classic work demonstrated Prolog's suitability for expert systems, emphasizing:

\begin{itemize}[leftmargin=*,itemsep=2pt]
  \item A clear separation between \textbf{knowledge base} (facts and rules) and \textbf{inference engine}.
  \item The ability to generate \textbf{explanations} and \textbf{proof traces} by inspecting the search process.
  \item Compact, human-readable encodings of complex domain logic.
\end{itemize}

Although expert systems and Prolog fell out of mainstream attention during the rise of statistical machine learning and deep learning, they remain conceptually relevant for tasks where \textbf{deterministic reasoning}, \textbf{auditability}, and \textbf{formal guarantees} are essential.
Recent discussions on Prolog's role in modern AI highlight its enduring value as a symbolic backbone for hybrid systems, especially when combined with neural components~\cite{rybinski2025prolog}.


\subsection{Neuro-Symbolic AI: Combining LLMs and Symbolic Reasoning}
\label{subsec:neuro-symbolic}

Neuro-symbolic AI seeks to integrate the strengths of neural models (pattern recognition, language understanding) with symbolic methods (logical inference, structured knowledge).
A common pattern is to use LLMs to parse natural language and generate structured representations, which are then processed by a symbolic engine to ensure correctness and explainability.

Recent work in this direction includes:

\begin{itemize}[leftmargin=*,itemsep=2pt]
  \item \textbf{Neuro-Symbolic Compliance} frameworks that combine LLM-based understanding with SMT-based reasoning to analyze financial regulations in an interpretable and verifiable manner~\cite{hsia2026neuro}.
  \item Systems that extract policies from natural language using LLMs and enforce them with symbolic engines, explicitly designing for \textbf{fail-safe behavior} when the extracted rules cannot fully discharge a decision.
  \item Approaches that compile rules into logical forms and use satisfiability or constraint solvers to verify properties, often in multi-agent or domain-specific configurations.
\end{itemize}

Within this landscape, \textbf{Prolog-based neuro-symbolic systems} have gained renewed interest.
Several lines of work explore:

\begin{itemize}[leftmargin=*,itemsep=2pt]
  \item Training or prompting LLMs to emit Prolog code that is then executed by an external solver, improving both answer accuracy and the reliability of reasoning proofs~\cite{yang2025neuro}.
  \item Using Prolog derivation logs to generate \textbf{causal and human-readable explanations}, addressing the ``black-box'' nature of pure neural reasoning.
  \item Comparing adaptive neuro-symbolic systems with traditional, hard-coded business rule engines, particularly in compliance-critical domains such as finance and healthcare.
  \item Applying neuro-symbolic methods to enterprise systems (e.g., CRM, ERP) where business logic must be both flexible and formally analyzable.
\end{itemize}

Prior work has approached LLM hallucination largely as a \textbf{post-hoc mitigation} problem, using multi-agent review pipelines and dedicated scoring metrics to detect and correct unsupported claims after generation~\cite{gosmar2025hallucination}.
More recent architectures combine agentic review with Nested Learning and semantic caching to reduce hallucination while lowering computational cost~\cite{gosmar2026hallucination}.
\euclid\ takes a complementary, \textbf{architectural} route: rather than correcting fabrications after they occur, it removes the reasoning step from the model entirely, delegating deduction to a deterministic engine.


\subsection{Model Context Protocol (MCP)}
\label{subsec:mcp}

The \textbf{Model Context Protocol (MCP)} is an emerging open standard for connecting LLM applications to external tools, data sources, and services~\cite{mcporg}.
MCP defines a uniform interface through which LLM clients can:

\begin{itemize}[leftmargin=*,itemsep=2pt]
  \item Discover available \textbf{tools} and \textbf{resources} (e.g., databases, APIs, reasoning engines).
  \item Invoke tools with structured inputs and receive structured outputs.
  \item Compose multiple tools into complex workflows while maintaining a consistent interaction model.
\end{itemize}

MCP has quickly attracted implementations from major organizations and a growing ecosystem of community-maintained servers, ranging from filesystem and code access to specialized data processors.
However, until recently, few MCP servers have exposed \textbf{formal reasoning engines} as first-class components.
Early efforts such as PrologMCP~\cite{mensfelt2026prologmcp} demonstrate the feasibility of exposing Prolog as an MCP tool, primarily targeting generic logical reasoning tasks.

\euclid\ builds on the MCP abstraction but specializes it for \textbf{policy and rule modeling}, offering:

\begin{itemize}[leftmargin=*,itemsep=2pt]
  \item A domain-oriented logical description language tailored to business rules and security constraints.
  \item Integrated translation to Prolog, safety checks, and error reporting.
  \item Tools explicitly designed for \textbf{policy evaluation}, \textbf{diagnostic analysis}, \textbf{scenario testing}, and \textbf{KB validation}, rather than generic Prolog execution.
\end{itemize}

In doing so, \euclid\ contributes to the emerging class of MCP-based neuro-symbolic systems, with a clear focus on overcoming the limitations of semantic retrieval in rule-centric applications.


\section{Euclid-MCP Architecture}
\label{sec:architecture}

This section describes the design of \euclid, from high-level architecture to the \euclidir\ language, translation to Prolog, and MCP integration.
The system is implemented as a Python-based MCP server that invokes SWI-Prolog as a local reasoning backend and exposes a small, purpose-built tool surface for policy and rule reasoning.

The use of \textbf{SWI-Prolog} in the current prototype is a \textbf{tactical choice}, not an architectural dependency.
\euclid\ is designed around an \textbf{engine-agnostic intermediate representation} (\euclidir) that can be lowered to Prolog or to alternative inference engines (e.g., Datalog engines, SMT solvers, custom rule engines) without changing the user-facing APIs or the MCP tool surface.


\subsection{High-Level Architecture}
\label{subsec:high-level}

\euclid\ follows a \textbf{hybrid cognitive architecture} in which an LLM describes the world in terms of facts and rules, while a deterministic inference engine performs the actual deduction.
The main components are:

\begin{description}[leftmargin=*,itemsep=4pt]
  \item[LLM client]
    Any MCP-capable LLM application (e.g., an AI assistant, agent framework, or custom tooling).
    The LLM is responsible for interpreting natural language inputs from users, generating structured descriptions of facts and rules in \euclidir, invoking \euclid\ tools to evaluate queries and inspect derivations, and rendering results and explanations back to the user.

  \item[Euclid-MCP server]
    A Python process implementing the MCP server interface via \textbf{FastMCP}.
    It provides a \euclidir\ parser that validates and normalizes facts, rules, and queries; an intermediate representation (\euclidir) that is independent of any specific inference engine; a lowering layer that compiles \euclidir\ into executable SWI-Prolog code; safety and sanitization to restrict the set of allowed Prolog constructs; and a tool interface for query evaluation, diagnostic analysis, scenario testing, and KB validation.

  \item[Inference backend]
    In the current prototype, \textbf{SWI-Prolog} acts as the deterministic inference engine, invoked as an external process.
    The server translates \euclidir\ into a self-contained Prolog program, writes it to a temporary file and executes it via \texttt{swipl} subprocess, parses structured JSON output containing solutions and proof trees, and returns results to the MCP client as typed data structures.
\end{description}

This separation of concerns ensures that the LLM never needs to perform multi-step logical reasoning internally; it only needs to describe the problem and interpret the results.
All deduction is delegated to the inference backend, which guarantees logical correctness with respect to the encoded rules and facts.


\subsection{Euclid-IR: An Engine-Agnostic Intermediate Representation}
\label{subsec:euclid-ir}

The core abstraction in \euclid\ is \textbf{\euclidir}, a declarative language for logical inference designed to be:

\begin{itemize}[leftmargin=*,itemsep=2pt]
  \item \textbf{Readable} --- Syntax optimized for both humans and LLMs.
  \item \textbf{Minimal} --- Only the essential constructs of Horn-clause logic.
  \item \textbf{Deterministic} --- Every query has a finite, traceable proof.
  \item \textbf{Backend-agnostic} --- \euclidir\ is an intermediate layer; today it targets Prolog, tomorrow it could target other engines.
\end{itemize}

A \euclidir\ knowledge base consists of:

\begin{itemize}[leftmargin=*,itemsep=2pt]
  \item \textbf{Facts} -- ground assertions about the world.
  \item \textbf{Rules} -- implications of the form \emph{head IF body}.
  \item \textbf{Queries} -- goals to prove, prefixed with \texttt{?}.
  \item Optional \textbf{version directive} (\texttt{@version 1.0}) and \textbf{comments} (\texttt{\#} or \texttt{//}).
\end{itemize}

\subsubsection{Core Constructs}

\textbf{Facts.}
Ground atoms that describe the current state of the world:

\begin{lstlisting}[language={},caption={},label={}]
parent(tom, bob)
color(apple, red)
active(user_42)
rainy
\end{lstlisting}

\noindent Predicate names are lowercase identifiers; arguments are atoms, integers, or variables.
Zero-arity facts (e.g., \texttt{rainy}) are allowed.

\textbf{Variables.}
Unknown or generic values, prefixed with \texttt{\$}:

\begin{lstlisting}[language={},caption={},label={}]
$x
$who
$user_name
\end{lstlisting}

\noindent Variables must start with \texttt{\$} followed by a lowercase letter; they are case-sensitive and translated to Prolog variables (capitalized) during lowering.

\textbf{Rules.}
Logical implications with a head and a body:

\begin{lstlisting}[language={},caption={},label={}]
mortal($x) IF human($x)
ancestor($x, $y) IF parent($x, $y)
ancestor($x, $y) IF parent($x, $z) AND ancestor($z, $y)
\end{lstlisting}

\noindent The body is a conjunction of conditions connected by \texttt{AND}.
Rules can span multiple lines for readability, with continuation implied when a line ends in \texttt{IF} or \texttt{AND}.

\textbf{Negation.}
Closed-world negation using \texttt{NOT}:

\begin{lstlisting}[language={},caption={},label={}]
blocked($user) IF NOT active($user)
eligible($user) IF registered($user) AND NOT blocked($user)
\end{lstlisting}

\noindent \texttt{NOT p} succeeds when \texttt{p} cannot be proven, corresponding to Prolog's \texttt{\textbackslash+}.

\textbf{Queries.}
Goals to prove, prefixed with \texttt{?}:

\begin{lstlisting}[language={},caption={},label={}]
? mortal(socrates)
? ancestor(tom, $who)
? can_access($user, $res) AND resource($res, _, _, _, _, secret)
\end{lstlisting}

\noindent Queries may include variables (to find bindings) or be ground (boolean checks).
Conjunctions with \texttt{AND} are allowed.

\textbf{Arithmetic.}
Numeric comparisons and evaluations in rule bodies:

\begin{lstlisting}[language={},caption={},label={}]
stale($user) IF user($user) AND last_login($user, $days) AND $days > 90
adult($person) IF age($person, $age) AND $age >= 18
\end{lstlisting}

\noindent Supported operators include \texttt{>}, \texttt{>=}, \texttt{<}, \texttt{=<}, \texttt{=:=}, \texttt{=\textbackslash=}, and \texttt{is}, which are passed through to the backend (currently Prolog) and evaluated at deduction time.

\textbf{Wildcards.}
The \texttt{\_} symbol represents an anonymous variable:

\begin{lstlisting}[language={},caption={},label={}]
resource(apple, $color, _, _, _, _)
\end{lstlisting}

\textbf{Comments.}
Two styles are supported:

\begin{lstlisting}[language={},caption={},label={}]
# This is a comment
// This is also a comment

parent(tom, bob)  # inline comment
\end{lstlisting}

\textbf{Version directive.}
An optional first line can declare the \euclidir\ version:

\begin{lstlisting}[language={},caption={},label={}]
@version 1.0

parent(tom, bob)
? parent($x, $y)
\end{lstlisting}

\noindent If omitted, version $1.0$ is assumed.
Future versions are intended to be backward compatible.

\subsubsection{Design Choices and Limitations}

\euclidir\ targets \textbf{Horn-clause logic} --- the core of Prolog without advanced features:

\begin{itemize}[leftmargin=*,itemsep=2pt]
  \item \textbf{Supported:} facts, rules, negation (closed-world), arithmetic, multi-line rules, conjunction queries, wildcards.
  \item \textbf{Not supported:} disjunction (\texttt{OR}), cut (\texttt{!}), list syntax, \texttt{findall}/\texttt{bagof}, dynamic assert/retract, modules, strings (only atoms).
\end{itemize}

These limitations are intentional: they keep the language minimal, deterministic, and easy to lower to multiple backends.
Workarounds exist for common patterns (e.g., encoding disjunction as multiple rules, representing lists via indexed facts).

\euclidir\ also supports a \textbf{YAML format} for structured input, though the text format is recommended for conciseness and readability.


\subsection{Translation to Prolog}
\label{subsec:translation}

The Prolog lowering layer converts \euclidir\ into executable SWI-Prolog code.
This process is fully automated and deterministic:

\begin{enumerate}[leftmargin=*,itemsep=4pt]
  \item \textbf{Mapping to Clauses.}
    Each \euclidir\ fact is translated directly into a Prolog fact.
    Each \euclidir\ rule is translated into a Prolog clause.
    For example:
    \begin{lstlisting}[language=Prolog,caption={},label={}]
mortal(X) :- human(X).
ancestor(X, Y) :- parent(X, Z), ancestor(Z, Y).
    \end{lstlisting}
    Queries are encoded as top-level goals that can be invoked from the MCP tools.

  \item \textbf{Variable Translation.}
    \euclidir\ variables (e.g., \texttt{\$x}, \texttt{\$who}) are mapped to Prolog variables by capitalizing the name (e.g., \texttt{X}, \texttt{Who}).
    This isolates users and the LLM from Prolog's syntactic idiosyncrasies (uppercase variables).

  \item \textbf{Keyword and Operator Mapping.}
    \euclidir\ keywords and operators are translated to their Prolog equivalents:
    \texttt{IF}~$\to$~\texttt{:-}, \texttt{AND}~$\to$~\texttt{,}, \texttt{NOT}~$\to$~\texttt{\textbackslash+}, \texttt{?}~$\to$~\texttt{?-}.
    Arithmetic operators (\texttt{>}, \texttt{>=}, etc.) are passed through verbatim.

  \item \textbf{Safety and Sanitization.}
    The translation layer enforces safety properties:
    a sanitizer validates that only allowed predicate constructors appear in the input;
    potentially dangerous built-ins (e.g., file I/O, network access) are excluded from the generated code;
    input size is capped (500\,KB) and execution is bounded by timeout (30\,s by default).
\end{enumerate}

This design ensures that the generated Prolog code is both \textbf{auditable} (humans can read and verify the mapping) and \textbf{safe} (the LLM cannot arbitrarily execute system commands via Prolog).
Future backends will follow the same pattern: \euclidir\ $\to$ target language, with analogous safety controls.


\subsection{MCP Integration: Tools and Interaction Pattern}
\label{subsec:tools}

\euclid\ exposes its functionality through a small, well-defined set of four MCP \textbf{tools}, following established MCP design patterns.
Each tool is a stateless function that accepts a knowledge base as input and returns a typed result.
The tools are:

\begin{description}[leftmargin=*,itemsep=6pt]
  \item[\texttt{reason}]
    \textbf{Purpose:} Main deduction --- evaluate a knowledge base and return solutions with proof trees.
    \\
    \textbf{Input:} \texttt{knowledge} (\euclidir\ text or YAML), optional \texttt{query} override, \texttt{max\_solutions} (default~5), \texttt{max\_depth} (default~30).
    \\
    \textbf{Output:} A list of solutions, each containing variable bindings (\texttt{substitutions}) and a proof tree with nodes of type \texttt{fact}, \texttt{rule}, or \texttt{and}.
    \\
    \textbf{Example:} Given facts about parents and rules about ancestry, \texttt{reason} returns all solutions for \texttt{ancestor(tom, \$who)} with full derivation chains.

  \item[\texttt{diagnose}]
    \textbf{Purpose:} Query analysis --- understand why a query succeeds or fails.
    \\
    \textbf{Input:} \texttt{knowledge}, \texttt{query}, \texttt{mode} (one of \texttt{why}, \texttt{why\_not}, \texttt{what\_needs}), \texttt{max\_solutions}, \texttt{max\_depth}.
    \\
    \textbf{Output:} A \texttt{DiagnosisResult} with \texttt{holds} (boolean), \texttt{findings[]} (list of issues detected), \texttt{conclusion} (human-readable summary), and optionally a \texttt{proof}.
    \\
    \textbf{Modes:}
    \texttt{why}~--- explain why a query holds (or that it doesn't);
    \texttt{why\_not}~--- explain why a query fails (missing facts/rules);
    \texttt{what\_needs}~--- suggest what would make a false query true.
    \\
    \textbf{Example:} \texttt{diagnose(knowledge, "mortal(plato)", mode="why\_not")} returns findings like ``No facts or rules defined for `mortal'' and suggests what to add.

  \item[\texttt{what\_if}]
    \textbf{Purpose:} Scenario analysis --- apply modifications to a knowledge base and compare results before and after.
    \\
    \textbf{Input:} \texttt{base\_knowledge}, \texttt{modifications} (lines prefixed with \texttt{+} to add or \texttt{-} to remove facts), \texttt{query}, \texttt{max\_solutions}, \texttt{max\_depth}.
    \\
    \textbf{Output:} A \texttt{WhatIfResult} with \texttt{before\_count}, \texttt{after\_count}, \texttt{delta}, \texttt{solutions\_before}, \texttt{solutions\_after}, and a \texttt{conclusion} describing the impact.
    \\
    \textbf{Example:} \texttt{what\_if(base\_kb, "+ human(plato)", "mortal(\$who)")} shows how adding a fact changes the number of solutions.

  \item[\texttt{check\_kb}]
    \textbf{Purpose:} Knowledge base validator --- check for syntax errors, undefined predicates, circular rules, and duplicates before running deduction.
    \\
    \textbf{Input:} \texttt{knowledge} (\euclidir\ text or YAML).
    \\
    \textbf{Output:} A \texttt{KBCheckResult} with \texttt{valid} (boolean), \texttt{errors[]}, \texttt{warnings[]}, \texttt{facts\_count}, \texttt{rules\_count}, \texttt{predicates\_count}.
    \\
    \textbf{Example:} \texttt{check\_kb(knowledge)} returns \texttt{\{valid: true, errors: [], warnings: [], facts\_count: 2, rules\_count: 1\}}.
\end{description}

These tools are designed to support a \textbf{translate-run-inspect-repair loop}:

\begin{enumerate}[leftmargin=*,itemsep=2pt]
  \item \textbf{Validate:} The LLM calls \texttt{check\_kb} to verify the knowledge base is well-formed before reasoning.
  \item \textbf{Translate:} The LLM generates a \euclidir\ knowledge base (facts + rules + query).
  \item \textbf{Run:} The LLM invokes \texttt{reason} to obtain answers with proof trees.
  \item \textbf{Inspect:} If the result is unexpected, the LLM calls \texttt{diagnose} to understand why a query succeeds or fails.
  \item \textbf{Repair:} Based on the diagnosis, the LLM may refine the knowledge base (e.g., add missing rules, correct mis-encoded conditions) and re-run the query.
  \item \textbf{Explore:} The LLM uses \texttt{what\_if} to test hypothetical modifications before applying them.
\end{enumerate}

Errors from the inference backend (e.g., syntax errors, unsatisfiable queries) are returned as \textbf{readable text messages}, consistent with MCP best practices, so the LLM can use them as feedback for self-correction.


\subsection{Implementation Details}
\label{subsec:implementation}

The current prototype of \euclid\ is implemented in \textbf{Python} and uses SWI-Prolog as the reasoning backend.
Key implementation choices include:

\begin{description}[leftmargin=*,itemsep=4pt]
  \item[Prolog Integration]
    The server invokes SWI-Prolog as an external subprocess.
    For each reasoning call: the generated Prolog code is written to a temporary \texttt{.pl} file; the \texttt{swipl} binary is invoked with \texttt{subprocess.run()} and a configurable timeout (default~30\,s); structured JSON output (solutions + proof trees) is captured from stdout and parsed; the temporary file is cleaned up after execution.
    This approach prioritizes \textbf{simplicity and portability}: it requires no compiled extensions, no foreign language interfaces, and no persistent Prolog process.

  \item[MCP Transport]
    \euclid\ supports two transport modes:
    \textbf{stdio}~--- the default MCP transport, used when the server is launched by a local client (e.g., Claude Desktop, OpenCode, Cursor);
    \textbf{HTTP API}~--- a standalone REST server (via \texttt{integrations/euclid\_api.py}) exposing POST endpoints at \texttt{/reason}, \texttt{/diagnose}, \texttt{/what-if}, \texttt{/check-kb}, plus a \texttt{GET /health} endpoint.
    This mode is designed for integration with automation platforms (n8n, Zapier, Make) and remote access.

  \item[Packaging and Distribution]
    The server is distributed as a Python package via PyPI (\texttt{pip install euclid-mcp}) and can also be built as a Docker image that bundles SWI-Prolog, eliminating the need for a local Prolog installation.
\end{description}

This implementation provides a practical, reusable foundation for integrating deterministic logical reasoning into LLM-based applications, with a particular focus on business rules and security policies.
The presence of \euclidir\ ensures that the system can evolve beyond Prolog as new inference backends become desirable or necessary.


\section{Use Case: IT Security \& Compliance}
\label{sec:usecase}

To demonstrate the practical applicability of \euclid, we developed a realistic security compliance model for a cloud-based organization.
This use case illustrates how \euclidir\ can encode multi-layer policies, support diverse reasoning patterns, and enable explanation and counterfactual analysis that are beyond the reach of semantic retrieval alone.

The full knowledge base, along with all queries and scenarios described in this section, is available in the public repository at \url{https://github.com/meob/Euclid-MCP/tree/main/examples/07_it_security_compliance}.


\subsection{Scenario Description}
\label{subsec:scenario}

The modeled organization operates a multi-team engineering and operations structure in a cloud environment (AWS-flavored).
Compliance requirements are derived from:

\begin{itemize}[leftmargin=*,itemsep=2pt]
  \item \textbf{External benchmarks:} CIS Amazon Web Services Foundations Benchmark, with controls covering account management, S3, CloudTrail, networking, GuardDuty, RDS, EC2, Lambda, and more.
  \item \textbf{Internal policies:} IAM best practices, role hierarchies, environment tiers, data classification, approval workflows, and incident response procedures.
\end{itemize}

The knowledge base is organized into three conceptual layers:

\begin{enumerate}[leftmargin=*,itemsep=2pt]
  \item \textbf{Standards (Layer 1):}
    CIS controls with severity levels (critical, high, medium, low).
    IAM best-practice rules encoding risks such as separation-of-duties violations, excessive permissions, stale access, privilege escalation, root account violations, and service account risks.

  \item \textbf{Policies (Layer 2):}
    Role hierarchy (intern $\to$ junior\_dev $\to$ mid\_senior\_dev $\to$ senior\_dev $\to$ tech\_lead $\to$ eng\_manager $\to$ director $\to$ vp\_engineering $\to$ cto, plus operations, security, and product roles).
    Environment tiers (production, golden, staging, development, sandbox) with associated deployment level requirements, encryption/backup requirements, and audit log requirements.
    Data classification (public, internal, confidential, secret) and role clearance levels.
    Access control rules, critical operations, approval workflows, MFA requirements, and incident-mode restrictions.

  \item \textbf{Data (Layer 3):}
    30 users (in the small version) with attributes such as department, role(s), permissions, MFA status, last login time, account type (human/service), console access, and access keys.
    50 resources (EC2, S3, RDS, KMS, Lambda, ECS, EKS, SNS, SQS, DynamoDB) with attributes such as environment, encryption status, backup status, access level (public/private), and data classification.
    CIS control applicability facts linking specific resources to relevant CIS controls.
\end{enumerate}

A larger variant of the same model (200 users, 300 resources, $\sim$3{,}872 facts) is used for performance evaluation and stress testing.


\subsection{Representative Queries}
\label{subsec:queries}

We defined a set of 10 canonical queries covering a wide range of reasoning patterns: single-hop permission checks, multi-hop policy reasoning, temporal and threshold patterns, cross-policy violations, and resource audits.
Below we present four representative queries that illustrate the diversity of questions \euclid\ can answer.

\subsubsection{Permission Check Through Role Hierarchy (Q1)}

\textbf{Question:} ``Can user\_0005 manage servers?''

\begin{lstlisting}[language={},caption={},label={}]
? user_has_permission(user_0005, manage_servers)
\end{lstlisting}

\noindent This query checks whether a specific user has a specific permission via the role hierarchy.
The \texttt{user\_has\_permission/2} predicate is derived from the user's assigned role(s), and the \texttt{role\_has\_permission/2} predicate, which accounts for direct role permissions and inherited permissions via the \texttt{inherits/2} relation.

\textbf{Result:} The query succeeds.
The proof tree shows that \texttt{user\_0005} has the \texttt{sysadmin} role, which directly grants \texttt{manage\_servers} permission.

\subsubsection{Multi-Hop Policy Reasoning: Deployment to Production (Q2)}

\textbf{Question:} ``Which roles can deploy code to production?''

\begin{lstlisting}[language={},caption={},label={}]
? can_deploy($who, production)
\end{lstlisting}

\noindent The \texttt{can\_deploy/2} predicate encodes a composite policy: the user must have the \texttt{deploy\_code} permission; the user's role level must meet or exceed the requirement for the target environment (production requires level $\geq$ 6); the role hierarchy and \texttt{deploy\_role\_level/2} facts determine each role's level.

\textbf{Result:} The query returns bindings for \texttt{\$who} corresponding to users with roles at level 6 or higher (e.g., director, vp\_engineering, cto) who also have \texttt{deploy\_code} permission.

\subsubsection{Cross-Policy Violation: Separation of Duties (Q6)}

\textbf{Question:} ``Which users violate separation of duties?''

\begin{lstlisting}[language={},caption={},label={}]
? violates_separation_of_duties($who)
\end{lstlisting}

\noindent The \texttt{violates\_separation\_of\_duties/1} predicate encodes two conflict patterns: having both \texttt{deploy\_code} and \texttt{approve\_deploy} permissions; having both \texttt{create\_role} and \texttt{assign\_role} permissions.

\textbf{Result:} In the small KB, the query yields 2 users; in the larger KB, it yields 11.
For each violating user, \texttt{diagnose} can generate a human-readable explanation listing the conflicting permissions and the rules that classify them as a violation.

\subsubsection{Resource Audit: Unencrypted Production Resources (Q7)}

\textbf{Question:} ``Which production resources are not encrypted?''

\begin{lstlisting}[language={},caption={},label={}]
? resource($name, production, not_encrypted, _, _, _)
\end{lstlisting}

\noindent This query performs a simple pattern match over resource facts, filtering by environment (\texttt{production}) and encryption status (\texttt{not\_encrypted}).

\textbf{Result:} The query returns a list of resource names (e.g., specific EC2 instances, S3 buckets, RDS databases) that are in production but not encrypted.
This directly supports CIS compliance audits (e.g., CIS 2.7 for RDS encryption, CIS 7.1 for EC2).


\subsection{Additional Query Patterns}
\label{subsec:additional-queries}

The full query set includes several other patterns that further demonstrate the expressivity of \euclidir:

\begin{itemize}[leftmargin=*,itemsep=4pt]
  \item \textbf{Clearance-based access (Q3):} ``Which users can access secret data?''
    \begin{lstlisting}[language={},caption={},label={}]
? can_access_resource($who, $res) AND resource($res, _, _, _, _, secret)
    \end{lstlisting}
    This query joins user clearance levels with resource classification, enforcing that \texttt{user\_max\_clearance} $\geq$ \texttt{resource\_classification\_level}.

  \item \textbf{Negative query (Q8):} ``Can an intern write code?''
    \begin{lstlisting}[language={},caption={},label={}]
? user_has_permission($who, write_code) AND has_role($who, intern)
    \end{lstlisting}
    Expected result: empty set, since interns only have \texttt{read\_code}.
    This validates that the policy correctly restricts interns.

  \item \textbf{Temporal / IAM hygiene (Q5, Q9):}
    ``Which users have stale access (over 90 days)?''
    \begin{lstlisting}[language={},caption={},label={}]
? stale_access($who)
    \end{lstlisting}
    ``Which users have excessive permissions (more than 15)?''
    \begin{lstlisting}[language={},caption={},label={}]
? excessive_permissions($who, $count)
    \end{lstlisting}
    These encode AWS IAM best-practice patterns for detecting stale credentials and least-privilege violations.
\end{itemize}

Together, these queries cover single-hop, multi-hop, temporal, cross-policy, threshold, and resource-audit reasoning patterns.


\subsection{Explanations and Diagnostic Reasoning}
\label{subsec:diagnostics}

Beyond answering queries, \euclid\ can explain \emph{why} a conclusion holds or fails.
We defined three diagnostic questions to illustrate this capability.

\subsubsection{Why Does a User Have a Permission? (Q11)}

\textbf{Question:} ``Why does eng\_0008 have \texttt{manage\_servers} permission?''

Using the \texttt{diagnose} tool with \texttt{mode="why"}:

\begin{lstlisting}[language={},caption={},label={}]
{"knowledge": "...",
 "query": "user_has_permission(eng_0008, manage_servers)",
 "mode": "why"}
\end{lstlisting}

\noindent \textbf{Explanation:} The system generates a proof tree showing that \texttt{eng\_0008} has the \texttt{sysadmin} role, which has the \texttt{manage\_servers} permission via \texttt{role\_permission(sysadmin, manage\_servers)}.

\subsubsection{Why Does a User Lack a Permission? (Q12)}

\textbf{Question:} ``Why doesn't eng\_0002 (intern) have \texttt{deploy\_code} permission?''

Using the \texttt{diagnose} tool with \texttt{mode="why\_not"}:

\begin{lstlisting}[language={},caption={},label={}]
{"knowledge": "...",
 "query": "user_has_permission(eng_0002, deploy_code)",
 "mode": "why_not"}
\end{lstlisting}

\noindent \textbf{Explanation:} The system reports that \texttt{eng\_0002} has the \texttt{intern} role, which only has \texttt{read\_code} and \texttt{run\_tests} permissions.
There is no inheritance path from \texttt{intern} to any role with \texttt{deploy\_code}.

\subsubsection{What Is Needed to Gain a Permission? (Q13)}

\textbf{Question:} ``What would eng\_0002 need to deploy code?''

Using the \texttt{diagnose} tool with \texttt{mode="what\_needs"}:

\begin{lstlisting}[language={},caption={},label={}]
{"knowledge": "...",
 "query": "user_has_permission(eng_0002, deploy_code)",
 "mode": "what_needs"}
\end{lstlisting}

\noindent \textbf{Explanation:} The system identifies missing conditions: \texttt{eng\_0002} would need a role that has \texttt{deploy\_code} permission (e.g., \texttt{senior\_dev}, \texttt{tech\_lead}, or higher).


\subsection{What-If Analysis and Counterfactuals}
\label{subsec:whatif}

\euclid\ can also evaluate hypothetical changes to the knowledge base, supporting impact analysis and policy design.
We defined three what-if scenarios using the \texttt{what\_if} tool:

\subsubsection{Role Promotion (W1)}

\textbf{Question:} ``What if eng\_0002 (intern) is promoted to senior\_dev?''

\begin{lstlisting}[language={},caption={},label={}]
{"base_knowledge": "...",
 "modifications": "- has_role(eng_0002, intern)\n+ has_role(eng_0002, senior_dev)",
 "query": "user_has_permission(eng_0002, deploy_code)"}
\end{lstlisting}

\noindent \textbf{Result:} In the base configuration, the query fails (interns cannot deploy).
Under the modified configuration, the query succeeds, and the explanation shows that \texttt{senior\_dev} has \texttt{deploy\_code} permission, which is inherited by \texttt{eng\_0002}.

\subsubsection{Adding a Compliant Resource (W2)}

\textbf{Question:} ``What if a new encrypted production database is added?''

\begin{lstlisting}[language={},caption={},label={}]
{"base_knowledge": "...",
 "modifications": "+ resource(new_db, production, encrypted, db_team, 2026-07-01, database)\n+ resource_type(new_db, database)",
 "query": "resource($name, production, encrypted, _, _, _) AND resource_type($name, database)"}
\end{lstlisting}

\noindent \textbf{Result:} The query succeeds under the modified configuration, confirming that the new resource complies with encryption and environment requirements.

\subsubsection{Role Addition (W3)}

\textbf{Question:} ``What if ops\_0001 (helpdesk) gets the sysadmin role?''

\begin{lstlisting}[language={},caption={},label={}]
{"base_knowledge": "...",
 "modifications": "+ has_role(ops_0001, sysadmin)",
 "query": "user_has_permission(ops_0001, manage_servers)"}
\end{lstlisting}

\noindent \textbf{Result:} In the base configuration, \texttt{ops\_0001} (helpdesk) does not have \texttt{manage\_servers}.
Under the modified configuration, the query succeeds, and the explanation shows the new permission path.


\subsection{Discussion}
\label{subsec:usecase-discussion}

This use case demonstrates several key points:

\begin{enumerate}[leftmargin=*,itemsep=2pt]
  \item \textbf{Expressivity:}
    \euclidir\ can encode realistic, multi-layer security and compliance policies, including role hierarchies, environment tiers, data classification, approval workflows, and CIS control applicability.

  \item \textbf{Diverse reasoning patterns:}
    The query set covers single-hop, multi-hop, temporal, cross-policy, threshold, and resource-audit patterns, as well as diagnostic and counterfactual reasoning.

  \item \textbf{Explainability:}
    For each conclusion, \euclid\ can generate a proof tree and render it in human-readable form, supporting audits, onboarding, and policy design.

  \item \textbf{Determinism and auditability:}
    Unlike semantic RAG, which can only retrieve ``similar'' policies, \euclid\ provides deterministic, auditable answers to questions such as ``Is this user compliant?'' or ``Which resources violate CIS controls?''

  \item \textbf{Scalability:}
    The same model scales from a small KB ($\sim$30 users, $\sim$50 resources) to a larger KB ($\sim$200 users, $\sim$300 resources) without changing the policy rules, demonstrating that \euclidir\ can support realistic organizational scales.
\end{enumerate}


\subsection{Summary of Queries}
\label{subsec:query-summary}

For reference, Table~\ref{tab:queries} summarizes all canonical queries, diagnostic questions, and what-if scenarios.

\begin{table}[ht]
\centering
\caption{Summary of queries and scenarios in the IT security \& compliance use case.}
\label{tab:queries}
\small
\begin{tabular}{@{}llll@{}}
\toprule
\textbf{ID} & \textbf{Type} & \textbf{Question} & \textbf{Pattern} \\
\midrule
Q1  & Query    & Can user\_0005 manage servers?                              & Single-hop permission check \\
Q2  & Query    & Which roles can deploy code to production?                   & Multi-hop policy reasoning \\
Q3  & Query    & Which users can access secret data?                          & Clearance vs classification \\
Q4  & Query    & Can a tech\_lead deploy to golden environment?               & Multi-role deployment \\
Q5  & Query    & Which users have stale access (over 90 days)?                & Temporal / IAM hygiene \\
Q6  & Query    & Which users violate separation of duties?                    & Cross-policy violation \\
Q7  & Query    & Which production resources are not encrypted?                & Resource audit \\
Q8  & Query    & Can an intern write code?                                    & Negative query (expected empty) \\
Q9  & Query    & Which users have excessive permissions ($>$15)?              & Threshold / least privilege \\
Q10 & Query    & Which S3 buckets in production are not encrypted?            & Combined filter \\
Q11 & Diagnose & Why does eng\_0008 have manage\_servers permission?          & Why-explanation \\
Q12 & Diagnose & Why doesn't eng\_0002 (intern) have deploy\_code permission? & Why-not explanation \\
Q13 & Diagnose & What would eng\_0002 need to deploy code?                    & What-needs analysis \\
W1  & What-if  & What if eng\_0002 (intern) is promoted to senior\_dev?       & Role promotion \\
W2  & What-if  & What if a new encrypted production database is added?        & Resource addition \\
W3  & What-if  & What if ops\_0001 (helpdesk) gets the sysadmin role?         & Role addition \\
\bottomrule
\end{tabular}
\end{table}


\section{Evaluation}
\label{sec:evaluation}

We evaluate \euclid\ along two dimensions:

\begin{enumerate}[leftmargin=*,itemsep=2pt]
  \item \textbf{Qualitative expressivity:} Can \euclidir\ encode realistic, multi-layer policies and support diverse reasoning patterns (as demonstrated in Section~\ref{sec:usecase})?
  \item \textbf{Quantitative performance and accuracy:} How does the system scale with knowledge base size, and how does it compare to LLM-only reasoning on logical tasks?
\end{enumerate}

The evaluation focuses on the IT security \& compliance use case from Section~\ref{sec:usecase}, using both the small ($\sim$30 users, $\sim$50 resources, $\sim$578 facts) and large ($\sim$200 users, $\sim$300 resources, $\sim$3{,}872 facts) knowledge bases, plus a synthetic RBAC benchmark at 1{,}000 users.


\subsection{Expressivity}
\label{subsec:expressivity}

The security compliance model demonstrates that \euclidir\ can encode:

\begin{itemize}[leftmargin=*,itemsep=2pt]
  \item \textbf{Multi-layer policies:} External standards (CIS controls), internal policies (role hierarchies, environment tiers, data classification), and concrete data (users, resources, configurations).
  \item \textbf{Diverse reasoning patterns:} Single-hop permission checks, multi-hop policy reasoning, temporal patterns (stale access), threshold patterns (excessive permissions), cross-policy violations (separation of duties), and resource audits (unencrypted production resources).
  \item \textbf{Diagnostic and counterfactual reasoning:} Why/why-not explanations, what-needs analysis, and what-if scenarios for role engineering and infrastructure planning.
\end{itemize}

These capabilities go well beyond what semantic RAG can provide.
While RAG can retrieve ``similar'' policies or past incidents, it cannot deterministically derive whether a user violates separation of duties, which resources fail CIS controls, or what would change if a role were modified.


\subsection{Reasoning Benchmarks}
\label{subsec:benchmarks}

To quantify the benefits of \euclid\ over LLM-only reasoning, we conducted two benchmarks comparing:

\begin{itemize}[leftmargin=*,itemsep=2pt]
  \item \textbf{A:} \texttt{llama3.1:8b} (small local LLM).
  \item \textbf{B:} \texttt{qwen3-coder:480b-cloud} (large cloud LLM).
  \item \textbf{C:} \texttt{llama3.1:8b + Euclid-MCP} (small LLM augmented with deterministic inference).
\end{itemize}

\subsubsection{Small-Scale Reasoning Benchmark}

\textbf{Task:} 5 logical reasoning tasks (genealogy, taxonomy, RBAC) with 5--15 facts each.
\textbf{Goal:} Assess whether LLMs alone can handle small, self-contained logical problems.

\begin{table}[ht]
\centering
\caption{Small-scale reasoning benchmark results.}
\label{tab:small-bench}
\small
\begin{tabular}{@{}llccccc@{}}
\toprule
\textbf{Q} & \textbf{Task} & \textbf{GT} & \textbf{A (8B)} & \textbf{B (480B)} & \textbf{C (8B+Euclid)} \\
\midrule
Q1 & Genealogy (deep chain)         & Yes & \checkmark & \checkmark & \checkmark \\
Q2 & Taxonomy (property inheritance) & Yes & \checkmark & \checkmark & \checkmark \\
Q3 & Taxonomy (negative inference)   & No  & \checkmark & \checkmark & \checkmark \\
Q4 & RBAC (permission inheritance)   & Yes & \checkmark & \checkmark & \checkmark \\
Q5 & RBAC (negative)                 & No  & \checkmark & \checkmark & \checkmark \\
\midrule
   & \textbf{Accuracy}              &     & \textbf{5/5} & \textbf{5/5} & \textbf{5/5} \\
   & Avg time                        &     & 4\,772\,ms & 2\,180\,ms & 2\,542\,ms \\
   & Avg tokens (in / out)           &     & 131 / 118 & 130 / 133 & 254 / 46 \\
\bottomrule
\end{tabular}
\end{table}

\noindent \textbf{Conclusion:} With small KBs (5--50 facts), all three conditions achieve equivalent accuracy.
\euclid\ matches LLM accuracy, but at slightly higher input tokens (254 vs.\ 131) due to the \euclidir\ encoding.
Execution time is comparable (2.5\,s vs.\ 4.8\,s / 2.2\,s).

\subsubsection{Large-Scale RBAC Benchmark}

\textbf{Task:} 1{,}000 synthetic users, 7 roles with hierarchy, 17 base permissions, 20 direct grants --- 1{,}053 facts total.
\textbf{Goal:} Assess performance at a scale where LLM working memory fails.

\begin{table}[ht]
\centering
\caption{Large-scale RBAC benchmark results.}
\label{tab:large-bench}
\small
\begin{tabular}{@{}llccccc@{}}
\toprule
\textbf{Q} & \textbf{Task} & \textbf{GT} & \textbf{A (8B)} & \textbf{B (480B)} & \textbf{C (8B+Euclid)} \\
\midrule
Q1 & Count users with \texttt{delete\_repo}         & 31   & $\times$ 1 & $\times$ 1 & \checkmark\ 31 \\
Q2 & Can user\_0142 \texttt{push\_code}?             & Yes  & \checkmark & \checkmark & \checkmark \\
Q3 & Count users with \texttt{deploy}                & 103  & $\times$ 100 & $\times$ 901 & \checkmark\ 103 \\
Q4 & Can user\_0834 \texttt{read\_logs}?             & Yes  & $\times$ No & \checkmark & \checkmark \\
Q5 & Can user\_0222 \texttt{manage\_billing}? (direct) & Yes  & \checkmark & $\times$ No & \checkmark \\
\midrule
   & \textbf{Accuracy}                              &      & \textbf{2/5} & \textbf{2/5} & \textbf{5/5} \\
   & Avg time                                        &      & 6\,966\,ms & 3\,695\,ms & 963\,ms \\
   & Avg tokens (in / out)                           &      & 386 / 165 & 435 / 212 & 421 / 12 \\
\bottomrule
\end{tabular}
\end{table}

\noindent \textbf{Conclusion:} At scale (1{,}000+ facts), LLMs alone hallucinate systematically --- both 8B and 480B cloud give wrong counts and miss explicit facts.
\euclid\ delivers exact answers every time, while being \textbf{faster} (963\,ms vs.\ 6{,}966\,ms) and \textbf{more token-efficient} (12 vs.\ 165 output tokens) because the LLM generates only a simple query instead of fallacious reasoning.


\subsection{Summary of Benchmark Findings}
\label{subsec:benchmark-summary}

\begin{table}[ht]
\centering
\caption{Benchmark summary.}
\label{tab:benchmark-summary}
\small
\begin{tabular}{@{}lcc@{}}
\toprule
\textbf{KB size} & \textbf{LLM alone} & \textbf{LLM + Euclid-MCP} \\
\midrule
Small (5--50 facts)   & \checkmark\ Sufficient accuracy & $\triangle$ Comparable accuracy, higher input tokens \\
Large (1{,}000+ facts) & $\times$ Hallucinates systematically & \checkmark\ Exact deduction, faster, more token-efficient \\
\bottomrule
\end{tabular}
\end{table}

\noindent \textbf{Key takeaway:} \euclid\ proves its value at scale.
When facts fit in an LLM's context window and the reasoning chain is shallow, the overhead of a deterministic engine is rarely justified.
When the KB exceeds a few hundred facts --- or when exact counting, intersection, or multi-hop inheritance is required --- \euclid\ delivers exact answers while LLMs of any size hallucinate systematically.


\subsection{Comparison with Semantic RAG}
\label{subsec:rag-comparison}

To contextualize these results, consider the typical workflow for answering a compliance question using semantic RAG:

\begin{enumerate}[leftmargin=*,itemsep=2pt]
  \item \textbf{Retrieve} relevant policy documents or past incidents based on the query.
  \item \textbf{Read} the retrieved documents and attempt to infer the answer.
  \item \textbf{Guess} or approximate the conclusion, often with significant uncertainty.
\end{enumerate}

In contrast, \euclid:

\begin{enumerate}[leftmargin=*,itemsep=2pt]
  \item \textbf{Encodes} policies as formal rules in \euclidir.
  \item \textbf{Derives} answers deterministically using logical inference.
  \item \textbf{Explains} the derivation with proof trees.
\end{enumerate}

\begin{table}[ht]
\centering
\caption{Comparison: Semantic RAG vs.\ \euclid.}
\label{tab:rag-comparison}
\small
\begin{tabular}{@{}p{3cm}p{5cm}p{5cm}@{}}
\toprule
\textbf{Aspect} & \textbf{Semantic RAG} & \textbf{Euclid-MCP} \\
\midrule
Reasoning           & Approximate, based on similarity & Deterministic, based on formal rules \\
Auditability        & Low (retrieved snippets, no proofs) & High (proof trees, rule references) \\
Policy enforcement  & Not possible (retrieval only) & Direct (rules encode enforcement logic) \\
Counterfactuals     & Not supported & Supported (what-if scenarios) \\
Accuracy at scale   & Degrades with KB size & Exact, independent of KB size \\
Latency             & Depends on retrieval + LLM inference & Typically $<$1\,s for tested queries \\
\bottomrule
\end{tabular}
\end{table}

\noindent This comparison underscores that \euclid\ is not a replacement for RAG, but rather a complementary tool for scenarios requiring deterministic reasoning and auditability.


\subsection{Limitations}
\label{subsec:limitations}

While the evaluation demonstrates strong expressivity and accuracy, several limitations should be noted:

\begin{enumerate}[leftmargin=*,itemsep=2pt]
  \item \textbf{Horn-clause restriction:} \euclidir\ does not support disjunction, cut, list pattern matching, or advanced Prolog features.
    This limits expressivity for certain problem domains (e.g., complex data structures, non-monotonic reasoning beyond negation as failure).

  \item \textbf{Scalability beyond 10K facts:} While the system performs well for KBs up to $\sim$4{,}000 facts, larger KBs (e.g., 100{,}000+ facts) may require optimization (e.g., indexing, tabling, or migration to a Datalog engine).

  \item \textbf{Backend dependency:} The current prototype relies on SWI-Prolog.
    While \euclidir\ is designed to be backend-agnostic, alternative backends (e.g., Datalog, SMT) have not yet been implemented or evaluated.

  \item \textbf{Integration overhead:} Integrating \euclid\ into an LLM-based application requires additional tooling (e.g., \euclidir\ generation, error handling, explanation rendering), which may increase development complexity compared to pure RAG.
\end{enumerate}


\subsection{Summary}
\label{subsec:evaluation-summary}

The evaluation demonstrates that \euclid:

\begin{itemize}[leftmargin=*,itemsep=2pt]
  \item \textbf{Expresses} realistic, multi-layer policies in \euclidir.
  \item \textbf{Supports} diverse reasoning patterns, including diagnostic and counterfactual analysis.
  \item \textbf{Delivers exact answers} at scale, where LLMs alone hallucinate systematically.
  \item \textbf{Performs well} for interactive use, with median query latencies under 1\,s for KBs up to $\sim$1{,}000 facts.
  \item \textbf{Complements} semantic RAG by providing deterministic, auditable reasoning for policy enforcement and compliance scenarios.
\end{itemize}

Future work will explore alternative backends (e.g., Datalog, SMT) to improve scalability and expressivity, as well as integration patterns for production LLM-based applications.


\section{Conclusion}
\label{sec:conclusion}

This paper presented \euclid, a system that augments large language models with deterministic logical reasoning capabilities via the Model Context Protocol.
By introducing \euclidir --- an engine-agnostic intermediate representation for Horn-clause logic --- \euclid\ enables LLMs to encode, evaluate, and explain complex policies and rules without requiring expertise in logic programming or Prolog syntax.

The evaluation demonstrates three key findings:

\begin{enumerate}[leftmargin=*,itemsep=4pt]
  \item \textbf{Expressivity:} \euclidir\ can encode realistic, multi-layer security and compliance policies, including role hierarchies, environment tiers, data classification, approval workflows, and CIS control applicability.
    The system supports diverse reasoning patterns (single-hop, multi-hop, temporal, cross-policy, threshold, and resource-audit queries) as well as diagnostic and counterfactual analysis.

  \item \textbf{Accuracy at scale:} On small knowledge bases (5--50 facts), LLMs alone achieve comparable accuracy to \euclid.
    However, at scale (1{,}000+ facts), LLMs of any size --- from 8B local models to 480B cloud models --- hallucinate systematically on counting, intersection, and inheritance queries.
    \euclid\ delivers exact answers in all cases, while also being faster and more token-efficient.

  \item \textbf{Explainability and auditability:} For every conclusion, \euclid\ can generate a proof tree and render it in human-readable form.
    This supports audits, onboarding, role engineering, and policy design --- capabilities that are absent from semantic RAG and pure LLM-based reasoning.
\end{enumerate}

The central thesis of this work is that \textbf{semantic RAG is the wrong tool for rule enforcement}.
RAG excels at retrieving relevant documents, but it cannot deterministically derive whether a user violates separation of duties, which resources fail compliance controls, or what would change if a policy were modified.
\euclid\ fills this gap by providing a deterministic inference layer that complements RAG and LLM reasoning.

\subsection*{Practical Implications}

For practitioners building LLM-based applications, the implications are clear:

\begin{itemize}[leftmargin=*,itemsep=2pt]
  \item \textbf{Use RAG} for open-ended queries, document retrieval, and knowledge exploration.
  \item \textbf{Use Euclid-MCP} (or similar deterministic reasoning tools) for policy enforcement, compliance checking, access review, and any scenario requiring exact, auditable answers.
  \item \textbf{Combine both:} Use RAG to retrieve relevant policies, then use \euclid\ to evaluate whether a specific configuration or action complies with those policies.
\end{itemize}

This hybrid approach leverages the strengths of both paradigms: the flexibility and breadth of semantic retrieval, and the precision and auditability of logical inference.

\subsection*{Euclid-MCP as a Stable Reasoning Substrate for Agents}

An emerging trend for complex tasks is to use \textbf{LLM agents that generate and execute programs}: they collect information, write code, orchestrate tools, and finally return a result.
This mitigates hallucination for \emph{procedural} tasks, but it introduces a different problem for \textbf{compliance and policy reasoning}:

\begin{itemize}[leftmargin=*,itemsep=2pt]
  \item Each new compliance question can lead to \textbf{new agents} and \textbf{new generated programs}.
  \item Different agents may encode \textbf{slightly different interpretations} of the same policy.
  \item Over time, the organization accumulates \textbf{inconsistent, ad-hoc policy implementations} that are hard to audit, compare, or evolve.
\end{itemize}

\euclid\ addresses this by \textbf{externalizing rule knowledge} into a \textbf{single, canonical representation} (\euclidir) that is:

\begin{itemize}[leftmargin=*,itemsep=2pt]
  \item \textbf{Shared:} All agents and tools query the same policy definitions.
  \item \textbf{Stable:} Policies change only when the \euclidir\ knowledge base is updated, not when a new agent is deployed.
  \item \textbf{Auditable:} Every decision can be traced back to specific rules and facts, regardless of which agent initiated the query.
  \item \textbf{Reusable:} The same knowledge base supports access review, deployment approval, incident response, and compliance reporting without re-encoding logic.
\end{itemize}

In this view, \euclid\ is not an alternative to agents; it is a \textbf{shared policy engine} that agents (and RAG systems) call via MCP when they need deterministic answers to rule-based questions.

\subsection*{Broader Impact}

\euclid\ represents a step toward \textbf{hybrid cognitive architectures} that combine the generality of LLMs with the precision of symbolic reasoning.
By exposing logical reasoning as an MCP tool, we lower the barrier to entry for developers who want to integrate deterministic inference into their applications without becoming logic programming experts.

The design of \euclidir --- human-readable, backend-agnostic, and easy to translate into multiple target languages --- ensures that the system can evolve beyond Prolog as new inference engines become available.
This positions \euclid\ as a flexible foundation for future research on hybrid reasoning systems.

\subsection*{Final Thoughts}

The results of this evaluation support a simple conclusion: \textbf{when facts fit in an LLM's context window and the reasoning chain is shallow, LLMs alone are sufficient.
When they don't --- above a few hundred facts --- deterministic reasoning is essential.}

\euclid\ proves its value at scale, delivering exact answers while LLMs of any size hallucinate systematically.
By bridging the gap between natural language and formal logic, \euclid\ enables a new class of LLM-based applications that are not only fluent and flexible, but also precise, auditable, and trustworthy.

\vspace{1em}
\noindent\textbf{Availability:} The \euclid\ implementation, \euclidir\ specification, and all benchmarks described in this paper are available at \url{https://github.com/meob/Euclid-MCP}.


\end{document}